\documentclass[conference]{IEEEtran}
\IEEEoverridecommandlockouts
\usepackage{cite}
\usepackage{amsmath,amssymb,amsfonts}
\usepackage{algorithmic}
\usepackage{graphicx}
\usepackage{textcomp}
\usepackage{xcolor}

\usepackage{graphicx}
\usepackage{algorithm}
\usepackage{algorithmic}
\usepackage{amsthm}
\usepackage{amssymb}
\usepackage{multicol}
\usepackage{amsmath}

\usepackage{bbm,amssymb,amsmath,amsfonts,graphicx,fullpage,ifthen,twoopt,algorithm,algorithmic,color}

\usepackage{times}

\def\BibTeX{{\rm B\kern-.05em{\sc i\kern-.025em b}\kern-.08em
    T\kern-.1667em\lower.7ex\hbox{E}\kern-.125emX}}
\begin{document}

\title{Etat de l'art sur l'application des bandits multi-bras \\

\thanks{Identify applicable funding agency here. If none, delete this.}
}

\author{\IEEEauthorblockN{1\textsuperscript{st} Djallel Bouneffouf}
\IEEEauthorblockA{\textit{IBM Research}\\
Djallel.bouneffouf@ibm.com\\
New York, USA }
}

\maketitle

\begin{abstract}
Le domaine des bandits multi-bras connaît actuellement une renaissance, alors que de nouveaux paramètres de problèmes et des algorithmes motivés par diverses applications pratiques sont introduits, en s'ajoutant au problème classique des bandits. Cet article vise à fournir un examen complet des principaux développements récents dans de multiples applications réelles des bandits. Plus précisément, nous introduisons une taxonomie des applications communes basées sur le MAB et résumons l'état de l'art pour chacun de ces domaines. De plus, nous identifions les tendances actuelles importantes et fournissons de nouvelles perspectives concernant l'avenir de ce domaine en plein essor.
\end{abstract}


\section{Introduction}
De nombreuses applications pratiques nécessitent des problèmes de prise de décision séquentiels, où un agent doit choisir la meilleure action parmi plusieurs alternatives. Des exemples de telles applications incluent les essais cliniques~\cite{durand2018contextual}, systèmes de recommandation~\cite{bouneffouf2012hybrid, 
bouneffouf2012following,allesiardo2014neural,
bouneffouf2013situation,bouneffouf2012considering,
bouneffouf2012exploration,bouneffouf2013applying,
bouneffouf2013risk,bouneffouf2013role,
bouneffouf2013drars,bouneffouf2013improving,
bouneffouf2014contextual,
bouneffouf2013towards,bouneffouf2008role,
bouneffouf2013contextual,bouneffouf2013impact,
bouneffouf2014recommandation,
bouneffouf2015sampling,
bouneffouf2013hybrid,
bouneffouf2013mobile,
bouneffouf2016exponentiated,
bouneffouf2013evolution,
bouneffouf2013apprentissage,
bouneffouf2014context,
bouneffouf2014etude,
bouneffouf2014freshness,
allesiardo2014prise,
bouneffouf2011temporal,
bouneffouf2014r,
bouneffouf2013proposition,
bouneffouflearning,
bouneffouf2013logique,
bouneffouf2014ant,
bouneffouf2016contextual,
bouneffouf2016multi,
bouneffouf2013exponentiated,
bouneffouf2016theoretical,
bouneffouf2013temporal,
bouneffouf2013optimizing,
bouneffouf2016ensemble,
bouneffouf2017context,
bouneffouf2017bandit,
bouneffoufdrars,
lin2018adaptive,
bouneffouf2018nystrom,
balakrishnan2018using,
riemer2019scalable,
balakrishnan2019incorporating,
bouneffouf2018eigenspectrum,
lin2018adaptive,
bouneffouffollowing,
riemer2017generative,
lin2018contextual,
choromanska2019beyond,
bouneffouf2020survey,
bouneffouf2013drars,
djallelrisk,
upadhyay2018bandit,
liu2019automated,
bouneffouf2019optimal,
noothigattu2018interpretable,
yurochkin2019online,
lin2019reinforcement,
noothigattu2019teaching,
aggarwal2019can,
balakrishnan2019using,
mehta2019ai,
liu2020admm,
sharma2020data,
balakrishnan2020constrained,
lin2020story,
varshneyteaching,
bouneffouf2020hyper,
lin2020unified,
bouneffouf2016learning,
lin2020online,
ram2020solving,
bouneffouf2020online,
bouneffouf2020contextual,
bouneffouf2013location,
toutanova2014proceedings,
leung2012neural,
jin2014neural,
bouneffouf2020computing,
bouneffouf2020spectral,
bouneffoufonline,
gupta20162016,
bouneffouf2015sampling,
bouneffouf2012contextual,
bouneffoufbandit,
bouneffoufsurvey,
bouneffoufspectral,
bouneffouf2018online,
bouneffouftoward} et la détection d'anomalies~\cite{Ding:2019}. Dans certains cas, des informations secondaires ou un contexte sont associés à chaque action (par exemple, le profil d'un utilisateur), et le retour d'information, ou récompense, est limité à l'option choisie. Par exemple, dans les essais cliniques, le contexte est le dossier médical du patient (par exemple, état de santé, antécédents familiaux, etc.), les actions correspondent aux options de traitement comparées et la récompense représente le résultat du traitement proposé (par exemple, succès ou échec). Un aspect important affectant le succès à long terme dans de tels contextes est de trouver un bon compromis entre l'exploration (par exemple, essayer une nouveau traitement) et l'exploitation (choisir le traitement le plus connue à ce jour).

Ce compromis inhérent entre l'exploration et l'exploitation existe dans de nombreux problèmes de prise de décision séquentiels, et est traditionnellement formulé comme le problème des bandit, qui se présente comme suit: Étant donné $K$ actions possibles, ou "bras", chacun associé à une distribution de probabilité de récompense fixe mais inconnue ~\cite{LR85,UCB}, à chaque itération, un agent sélectionne un bras à jouer et reçoit une récompense, échantillonnée à partir de la distribution de probabilité du bras respectif indépendamment des actions précédentes. La tâche d'un agent est d'apprendre à choisir ses actions afin que les récompenses cumulées au fil du temps soient maximisées.

Notez que l'agent doit essayer différentes bras pour apprendre leurs récompenses (c'est-à-dire explorer le gain), et également utiliser ces informations apprises afin de recevoir le meilleur gain (exploiter les gains appris). Il existe un compromis naturel entre l'exploration et l'exploitation. Par exemple, essayer chaque bras exactement une fois, puis jouer le meilleur d'entre. Cette approche est souvent susceptible de conduire à des solutions très sous-optimales lorsque les récompenses des bras sont incertaines. Différentes solutions ont été proposées pour ce problème, basées sur une formulation stochastique ~\cite{LR85,UCB,BouneffoufF16} et une formulation bayésienne ~\cite{AgrawalG12}; cependant, ces approches ne tenaient pas compte du contexte ou des informations secondaires dont disposait l'agent.

Il est à noté que le problème des bandits peut être vu comme la forme la plus simple d'apprentissage par renforcement, dans laquelle l'agent est sans état. Lorsque le système a des états, les actions provoquent des changements d'états et les récompenses dépendent également des états. Par conséquent, dans l'apprentissage par renforcement, les récompenses à différentes étapes ne sont pas indépendantes les unes des autres. En fait, les algorithmes classiques pour l'apprentissage par renforcement (avec états) utilisent souvent des solutions au problème des bandits multi-bras comme sous-programmes pour définir des politiques exploration exploitation dans l'apprentissage par renforcement. Par exemple, il est bien connu $\epsilon$-greedy L'algorithme de bandits multi-bras est souvent combiné avec l'algorithme de programmation dynamique de Bellman pour l'apprentissage par renforcement afin de définir les choix d'actions. En outre, de nombreux algorithmes d'apprentissage par renforcement, lorsqu'ils sont appliqués à des systèmes sans état, se réduisent à des algorithmes de bandit multi-bras.   

Une version particulièrement utile du MAB est le contextual multi-arm bandit (CMAB), ou simplement le contextual bandit, où à chaque itération, avant de choisir un bras, l'agent observe un $N$-dimensions du contexte, ou vecteur de features.
L'agent utilise ce contexte, ainsi que les récompenses des bras jouées dans le passé, pour choisir quel bras jouer dans l'itération actuelle. Au fil du temps, le but de l'agent est de collecter suffisamment d'informations sur la relation entre les vecteurs de contexte et les récompenses, afin qu'il puisse prédire le prochain meilleur bras à jouer en regardant le contexte actuel \cite{langford2008epoch,AgrawalG13}. Différents algorithmes ont été proposés pour le cas général, dont LINUCB~\cite{Li2010}, Neural Bandit \cite{AllesiardoFB14} et Contextual Thompson Sampling (CTS)~\cite{AgrawalG13}, où une dépendance linéaire est généralement supposée entre la récompense attendue d'une action et son contexte. 

Nous allons maintenant fournir un aperçu des diverses applications du bandit au problèmes de la vie réelle (santé, réseau informatique, finance, et au-delà), ainsi qu'en apprentissage automatique. En particulier, lorsque les approches bandit peuvent aider à améliorer le réglage des hyperparamètres et d'autres choix algorithmiques importants dans l'apprentissage supervisé, l'apprentissage actif et l'apprentissage par renforcement. 

\section{Applications des Bandits }
Le bandit stochastique aborde les défis associés à la présence d'incertitude dans la prise de décision séquentielle. Ce type d'incertitude a une interaction complexe avec le dilemme de l'exploration exploitation et fournit donc un formalisme naturel pour la plupart des problèmes de prise de décision.

\subsection{Santé}
\textbf {Essais cliniques. } La collecte de données pour évaluer l'efficacité du traitement sur des animaux pendant tous les stades de la maladie peut être difficile lors de l'utilisation de procédures conventionnelles d'allocation de traitement aléatoire, car de mauvais choix de traitements peuvent entraîner une détérioration de la santé du sujet. Les auteurs de \cite{durand2018contextual} visent à concevoir une stratégie d'allocation adaptative pour améliorer l'efficacité de la collecte de données en allouant plus d'échantillons pour explorer des traitements prometteurs. Ils présentent cette application comme un problème de bandit contextuel et introduisent un algorithme pratique d'exploration exploitation dans ce cadre. Le travail repose sur le sous-échantillonnage pour comparer les options de traitement en utilisant une quantité équivalente d'informations. Ils étendent la stratégie de sous-échantillonnage au contexte de bandit contextuel en appliquant un sous-échantillonnage dans la régression avec processus gaussien. 

Warfarine est l'anticoagulant oral le plus utilisé dans le monde; cependant, l'administration d'un dosage précis reste un défi important, car le dosage approprié peut être très variable entre les individus en raison de divers facteurs cliniques, démographiques et génétiques. Les médecins suivent actuellement une stratégie à dose fixe: les patients commencent avec une dose de 5 mg / jour (ce qui est la posologie appropriée pour la majorité des patients) et ajustent lentement la dose au cours de quelques semaines en suivant les taux d’anticoagulant du patient. Cependant, une posologie initiale incorrecte peut entraîner des conséquences très néfastes telles qu'un accident vasculaire cérébral (si la dose initiale est trop faible) ou une hémorragie interne (si la dose initiale est trop élevée). Ainsi, les auteurs de \cite{bastani2015online} abordent le problème de l'apprentissage et de l'attribution d'un dosage initial approprié aux patients en modélisant le problème comme un bandit avec des covariables de haute dimension, et proposent un nouvel algorithme de bandit efficace basé sur l'estimateur LASSO.

\textbf{Modélisation du cerveau et du comportement.} S'inspirant des études comportementales de la prise de décision humaine chez les patients souffrant de différents troubles mentaux, les auteurs de \cite{bouneffouf2017bandit} proposent un cadre paramétrique général pour le problème des bandits qui étend l'approche standard d'échantillonnage de Thompson pour incorporer les biais de traitement des récompenses associés à plusieurs conditions neurologiques et psychiatriques, y compris les maladies de Parkinson et d'Alzheimer, le trouble de déficit de l'attention/ hyperactivité (TDAH), la dépendance et la douleur chronique. Ils démontrent empiriquement, du point de vue de la modélisation comportementale, que leur model peut être considéré comme une première étape vers un modèle de calcul unificateur capturant les anomalies du traitement des récompenses dans plusieurs conditions mentales. 

\subsection{La finance}
Ces dernières années, la sélection séquentielle de portefeuilles a suscité un intérêt croissant à l'intersection de l'apprentissage automatique et de la finance quantitative. Le compromis entre l'exploration et l'exploitation, dans le but de maximiser la récompense cumulative, est une formulation naturelle des problèmes de choix de portefeuille. Dans \cite{shen2015portfolio}, les auteurs ont proposé un algorithme de bandit pour faire des choix de portefeuille en ligne en exploitant les corrélations entre plusieurs bras. En construisant des portefeuilles orthogonaux à partir de plusieurs actifs et en intégrant leur approche au cadre des bandits, les auteurs dérivent la stratégie de portefeuille optimale représentant une combinaison d'investissements passifs et actifs selon une fonction de récompense ajustée au risque. 
Dans \cite{huo2017risk}, les auteurs intègrent la conscience du risque dans le cadre classique du bandit et introduisent un nouvel algorithme pour la construction de portefeuille. En filtrant les actifs en fonction de la structure topologique du marché financier et en combinant la politique optimale de bandit avec la minimisation d'une mesure de risque, ils parviennent à un équilibre entre le risque et le rendement.

\subsection{Tarification dynamique}
Les entreprises de vente en ligne sont souvent confrontées au problème de tarification dynamique: l'entreprise doit décider des prix en temps réel pour chacun de ses multiples produits. L'entreprise peut mener des expériences de prix (faire des changements de prix fréquents) pour se renseigner sur la demande et maximiser les profits à long terme. Les auteurs de \cite{misra2018dynamic} proposent une politique d'expérimentation dynamique des prix, où l'entreprise ne dispose que d'informations incomplètes sur la demande. Pour ce paramètre général, les auteurs dérivent un algorithme de tarification qui équilibre le fait de gagner un profit immédiat par rapport à l'apprentissage pour les bénéfices futurs. L'approche combine un bandit avec une identification partielle de la demande des consommateurs à partir de la théorie économique. Semblable à \cite{misra2018dynamic}, les auteurs de \cite{mueller2018low} considèrent la tarification multi-produits dynamique de haute dimension avec un modèle de demande linéaire de faible dimension évolutif. Ils montrent que le problème de maximisation des revenus se réduit à une optimisation convexe de bandit en ligne avec des informations secondaires données par les demandes observées. L'approche applique un algorithme d'optimisation convexe de bandit dans un espace projeté de faible dimension couvert par les caractéristiques du produit latent, tout en apprenant simultanément cette durée via la décomposition en valeur singulière en ligne d'une matrice contenant les demandes observées.
 
\subsection{Systèmes de recommandation}
Les systèmes de recommandation sont fréquemment utilisés dans diverses applications pour prédire les préférences de l'utilisateur. Cependant, ils sont également confrontés au dilemme exploration-exploitation lorsqu'ils font une recommandation, car ils doivent exploiter leurs connaissances sur les éléments précédemment choisis qui intéressent l'utilisateur, tout en explorant de nouveaux éléments susceptibles de plaire à l'utilisateur. Les auteurs de \cite{zhou2017large} abordent ce défi en utilisant le paramètre bandit, en particulier pour les systèmes de recommandation à grande échelle qui ont un nombre vraiment grand ou infini d'éléments. Ils proposent deux approches de bandit à grande échelle dans des situations où aucune information préalable n'est disponible. Une exploration continue de leurs approches peut résoudre le problème du démarrage à froid dans les systèmes de recommandation. Dans les systèmes de recommandation contextuels, la plupart des approches existantes se concentrent sur la recommandation d'éléments pertinents aux utilisateurs, en tenant compte des informations contextuelles, telles que l'heure, le lieu ou les aspects sociaux. Cependant, aucune de ces approches n’a pris en compte le problème de l’évolution du contenu des utilisateurs. Dans \cite{bouneffouf2012contextual}, les auteurs introduisent un algorithme qui prend en compte cette dynamique. Il est basé sur une exploration / exploitation dynamique et peut équilibrer de manière adaptative les deux aspects, en décidant quelle situation est la plus pertinente pour l'exploration ou l'exploitation.
En ce sens, \cite{bouneffouf2014freshness} propose d'étudier la "fraîcheur" du contenu de l'utilisateur à travers le problème du bandit. Ils introduisent l'algorithme Freshness-Aware Thompson Sampling pour la recommandation de nouveaux documents.

\subsection{Maximisation de l'influence}
Les auteurs de \cite{vaswani2017model} considèrent la maximisation de l'influence (IM) dans les réseaux sociaux, qui est le problème de maximiser le nombre d'utilisateurs qui prennent conscience d'un produit en sélectionnant un ensemble d'utilisateurs auxquels exposer le produit. Ils proposent une nouvelle paramétrisation qui rend non seulement le cadre indépendant du modèle de diffusion sous-jacent, mais aussi statistiquement efficace pour apprendre des données. 

Ils donnent une fonction de substitution monotone et submodulaire correspondante, et montrent qu'il s'agit d'une bonne approximation de l'objectif original de la MI. Ils considèrent également le cas d'un nouveau marketeur cherchant à exploiter un réseau social existant, tout en apprenant simultanément les facteurs régissant la propagation de l'information. Pour cela, ils développent un algorithme de bandit basé sur LinUCB. Les auteurs de \cite{wen2017online} étudient également le problème de maximisation de l'influence en ligne dans les réseaux sociaux mais sous le modèle de cascade indépendant. Plus précisément, ils essaient d'apprendre l'ensemble des "meilleures graines ou influenceurs" dans un réseau social en ligne tout en interagissant à plusieurs reprises avec lui. Ils abordent les défis de l'espace d'action combinatoire, car le nombre d'ensembles d'influenceurs réalisables augmente de manière exponentielle avec le nombre maximum d'influenceurs et un retour limité, car seule la partie influencée du réseau est observée. 

\subsection{Récupération de l'information}
Les auteurs de \cite{losada2017multi} soutiennent que le processus de sélection itérative de recherche d'informations peut être naturellement modélisé comme un problème de bandit contextuel. Le modèle de bandit conduit à des méthodes très efficaces pour l'arbitrage des documents. Dans ce cadre d'attribution des bandits, ils proposent sept nouvelles méthodes de jugement de documents, dont cinq sont des méthodes stationnaires et deux sont des méthodes non stationnaires. Cette étude comparative comprend les méthodes existantes conçues pour l'évaluation basée sur la mise en commun et les méthodes existantes conçues pour la méta-recherche. Dans la recherche d'informations mobiles, les auteurs de \cite{bouneffouf2013contextual} introduisent un algorithme qui aborde ce dilemme dans le domaine de la recherche d'informations basées sur le contexte (CBIR). Il est basé sur une exploration / exploitation dynamique et il peut équilibrer de manière adaptative les deux aspects en décidant quelle situation d’utilisateur est la plus pertinente pour l’exploration ou l’exploitation. Dans un cadre en ligne délibérément conçu, ils effectuent des évaluations auprès des utilisateurs mobiles.


\subsection{Systèmes de Dialogue}
\textbf {Sélection de réponse de dialogue.} La sélection de réponse de dialogue est une étape importante vers la génération de réponse naturelle dans les agents conversationnels. Les travaux existants sur les modèles conversationnels se concentrent principalement sur l'apprentissage supervisé hors ligne à l'aide d'un large ensemble de paires contexte-réponse. Dans \cite{LiuYLM18}, les auteurs se concentrent sur l'apprentissage en ligne de la sélection des réponses dans les systèmes de dialogue. Ils proposent un modèle de bandit contextuel avec une fonction de récompense non linéaire qui utilise une représentation distribuée du texte pour la sélection de réponse en ligne. Un LSTM bidirectionnel est utilisé pour produire les représentations distribuées du contexte de dialogue et des réponses, qui servent d'entrée à un bandit contextuel. Ils proposent une méthode d'échantillonnage personnalisée de Thompson qui est appliquée à un espace de caractéristiques polynomiales pour approximer la récompense.

\textbf{Systèmes de dialogue.} L'objectif de la pro-activité dans les systèmes de dialogue est d'améliorer la convivialité des agents conversationnels en leur permettant d'initier des conversations. Alors que les systèmes de dialogue sont devenus de plus en plus populaires, les systèmes de dialogue actuels axés sur les tâches sont principalement réactifs, car les utilisateurs humains ont tendance à lancer des conversations. Les auteurs de \cite{silander2018contextual} proposent d'introduire le paradigme des bandits contextuels comme cadre pour des systèmes de dialogue proactifs. Les bandits contextuels ont été le modèle de choix pour le problème de la maximisation des récompenses avec rétroaction partielle car ils correspondent bien à la description de la tâche, ils explorent également la notion de mémoire dans ce paradigme, où ils proposent deux modèles de mémoire différentiables qui agissent comme des parties du fonction d'estimation de récompense paramétrique. Le premier, les réseaux de mémoire sélective par convolution, utilise une sélection d'interactions passées dans le cadre de l'aide à la décision. Le deuxième modèle, appelé réseau de mémoire attentive contextuelle, met en oeuvre un mécanisme d'attention différentiable sur les interactions passées de l'agent. Le but est de généraliser le modèle classique des bandits contextuels aux contextes où les informations temporelles doivent être incorporées et exploitées de manière apprenable.
 
\textbf{Systèmes de dialogue multi-domaines.} Construire des agents de dialogue multi-domaines est une tâche difficile et un problème ouvert dans l'IA moderne. Dans le domaine du dialogue, la capacité d'orchestrer plusieurs agents de dialogue formés indépendamment, ou compétences, pour créer un système unifié est d'une importance particulière. Dans \cite{upadhyaybandit}, les auteurs étudient la tâche d'orchestration du dialogue en ligne, où ils définissent l'orchestration postérieure comme la tâche de sélectionner un sous-ensemble de compétences qui répond le mieux à une entrée utilisateur en utilisant des fonctionnalités extraites à la fois de l'entrée utilisateur et de l'individu compétences. Pour tenir compte des coûts variés associés à l'extraction des caractéristiques des compétences, ils considèrent l'orchestration postérieure en ligne avec un budget d'exécution des compétences. Ce paramètre est formalisé en tant que bandit attentif au contexte avec observations, une variante des bandits attentifs au contexte, puis l'évalue sur des ensembles de données conversationnelles simulées.

\subsection{Détection d'une anomalie}
Les auteurs de \cite{Ding:2019} étudient le problème de la détection d'anomalies dans un cadre interactif. Leur objectif est de maximiser les véritables anomalies présentées à l'expert humain après épuisement d'un budget donné. Parallèlement à cette ligne, ils formulent le problème à travers le cadre de bandit et développent un nouvel algorithme de bandit contextuel collaboratif, qui modélise explicitement les attributs et les dépendances de noeuds de manière transparente dans un cadre commun, et gère le dilemme exploration-exploitation lors de l'interrogation. 

Les transactions par carte de crédit susceptibles d'être frauduleuses par les systèmes de détection automatisés sont généralement transmises à des experts humains pour vérification. Pour limiter les coûts, il est courant de ne sélectionner que les transactions les plus suspectes pour enquête. Les auteurs de \cite{soemers2018adapting} affirment qu'un compromis entre l'exploration et l'exploitation est impératif pour permettre l'adaptation aux changements de comportement. L'exploration consiste en la sélection et l'investigation des transactions dans le but d'améliorer les modèles prédictifs, et l'exploitation consiste à enquêter sur les transactions détectées comme suspectes. Modélisant la détection des transactions frauduleuses comme une récompense, ils utilisent un apprenant d'arbre de régression incrémentiel pour créer des grappes de transactions avec des récompenses attendues similaires. Cela permet l'utilisation d'un algorithme de bandit contextual(CMAB) pour fournir le compromis exploration / exploitation.

\subsection{ Télécommunication}
Dans \cite{boldrini2018mumab}, un modèle de bandit a été utilisé pour décrire le problème de la meilleure sélection de réseau sans fil par un dispositif multi-Radio Access Technology (multi-RAT), dans le but de maximiser la qualité perçue par l'utilisateur final. Le modèle proposé étendre le modèle MAB classique de deux manières. Premièrement, il prévoit deux actions différentes: mesurer et utiliser; deuxièmement, il permet aux actions de s'étaler sur plusieurs étapes de temps. Deux nouveaux algorithmes conçus pour tirer parti de la plus grande flexibilité offerte par le modèle muMAB ont également été introduits. Le premier, appelé mesure-utilisation-UCB1 est dérivé de l'algorithme UCB1, tandis que le second, appelé Mesure avec intervalle logarithmique, est conçu de manière appropriée pour le nouveau modèle afin de tirer parti de la nouvelle action de mesure, tout en en utilisant agressivement le meilleur bras.
Les auteurs de \cite{KerkoucheAFVM18} démontrent la possibilité d'optimiser les performances de la technologie Long Range Wide Area Network. Les auteurs suggèrent que les nœuds utilisent des algorithmes de bandit multi-bras, pour sélectionner les paramètres de communication (facteur d'étalement et puissance d'émission). Les évaluations montrent que de telles méthodes d'apprentissage permettent de gérer bien mieux le compromis entre la consommation d'énergie et la perte de paquets qu'un algorithme Adaptive Data Rate adaptant les facteurs d'étalement et les puissances de transmission sur la base des valeurs du rapport signal sur interférence et du rapport de bruit.

 \subsection{Bandit dans les applications réelles: résumé et orientations futures}
 
\begin{table}[h]
\scriptsize
\caption {Application des Bandits dans la vie réelle}
\label{tab:Life} 
\begin{tabular}{|l|r|l|l|l|}
\hline
                            &     & Non-   &       & Non-   \\ 
                            & MAB     & stat  &  CMAB      &  stat  \\
                            &      & MAB    &       & CMAB  \\
                            \hline
Santé        &  $\surd$ &                      &  $\surd$  &                                  \\ \hline
La finance           &   $\surd$ &                     &           &                                  \\ \hline
Tarification dynamique  &         &     $\surd$           &           &                                  \\ \hline
Système de recommandation & $\surd$  &    $\surd$           & $\surd$   &     $\surd$                          \\ \hline
Maximisation      & $\surd$  &                      &           &                                  \\ \hline
Système de dialogue   &          &                      &   $\surd$  &                                  \\ \hline
Télécomunication  &  $\surd$ &                      &           &                                  \\ \hline
Détection d'anomalie          &   $\surd$ &                      &           &                                  \\ \hline
\end{tabular}
\end{table}
Le tableau \ref{tab:Life} fournit un résumé des formulations de problèmes de bandit utilisées dans diverses applications spécifiques à un domaine. Le choix du modèle de bandit est souvent spécifique au domaine. Par exemple, il est évident que le bandit non stationnaire n'a pas été utilisé dans les applications de soins de santé, car des changements significatifs ne sont pas attendus dans le processus de prise des décisions de traitement, c'est-à-dire pas de transition dans l'état du patient; de telles transitions, si elles se produisaient, seraient mieux modélisées en utilisant l'apprentissage par renforcement plutôt que le bandit non stationnaire. Il existe clairement d'autres domaines où le bandit non stationnaire est un cadre plus approprié, mais il semble que ce paramètre n'ait pas encore été étudié de manière significative dans les domaines de la santé. Par exemple, la détection d'anomalie est un domaine dans lequel un bandit contextuel non stationnaire pourrait être utilisé, car dans ce contexte, l'anomalie pourrait être contradictoire, ce qui signifie que tout bandit appliqué à ce paramètre devrait avoir une sorte de condition de dérive, afin de s'adapter à de nouveaux types d'attaques. Un autre constat est qu'aucun des travaux existants n'a tenté de développer un algorithme capable de résoudre ces différentes tâches en même temps, ou d'appliquer les connaissances obtenues dans un domaine à un autre domaine, ouvrant ainsi une direction de recherche sur le multitask et le transfer learning dans le cadre de bandit. De plus, étant donné la nature en ligne du problème de bandit, le lifelong learning serait une prochaine étape naturelle.


\section{Bandit pour un meilleur apprentissage automatique}
Dans cette section, nous décrivons comment les algorithmes de bandit pourraient être utilisés pour améliorer d'autres algorithmes, par ex. diverses techniques d'apprentissage automatique.

\subsection{Selection d'algorithm }
La sélection de l'algorithme est généralement basée sur des modèles de performances d'algorithme, appris au cours d'une séquence d'apprentissage hors ligne distincte, qui peut être d'un coût prohibitif. Dans des travaux récents, ils ont adopté une approche en ligne, dans laquelle un modèle de performance est mis à jour de manière itérative et utilisé pour guider la sélection. Le compromis exploration-exploitation qui en résultait a été représenté comme un problème de bandit avec des conseils d'experts, en utilisant un solveur existant pour ce jeu, cela nécessitait l'utilisation d'une limite arbitraire sur les temps d'exécution de l'algorithme, annulant ainsi le regret optimal du solveur. Dans \cite{GaglioloS10}, un cadre plus simple a été proposé pour représenter la sélection d'algorithmes comme un problème de bandit, en utilisant des informations partielles et une limite inconnue sur les pertes.

\subsection{Optimisation des hyperparamètres}
\cite{li2016hyperband} a formulé l'optimisation des hyperparamètres comme un problème de bandit non stochastique d'exploration pure où des ressources prédéfinies, telles que des itérations, des échantillons de données ou des fonctionnalités sont allouées à des configurations échantillonnées aléatoirement. Ces travaux ont introduit un nouvel algorithme, Hyperband, pour ce cadre et analyse ses propriétés théoriques, offrant plusieurs garanties. En outre, Hyperband était comparé aux méthodes d'optimisation bayésiennes populaires; il a été observé qu'Hyperband peut fournir une accélération plus grande par rapport à ses concurrents sur une variété de problèmes d'apprentissage.

\subsection{Sélection des Features}
Dans un apprentissage supervisé en ligne classique, la véritable étiquette d'un échantillon est toujours révélée au classificateur, contrairement à un bandit où une mauvaise classification se traduit par une récompense nulle, et seule la classification correcte donne la récompense 1. Les auteurs de ~ \cite{wang2014online} étudie le problème de la sélection des fonctionnalités en ligne, où le but est de faire des prédictions précises en utilisant seulement un petit nombre de fonctionnalités actives à l'aide de l'algorithme epsilon-greedy. Les auteurs de \cite{BouneffoufRCF17} abordent le problème de la sélection de fonctionnalités en ligne en abordant le problème d'optimisation combinatoire dans le cadre de bandit stochastique avec retour de bandit, en utilisant l'algorithme d'échantillonnage de Thompson. 

\subsection{Bandit pour l'apprentissage actif}
L'étiquetage de tous les exemples dans un cadre de classification supervisée peut être coûteux. Les stratégies d'apprentissage actif résolvent ce problème en sélectionnant les exemples non étiquetés les plus utiles pour obtenir l'étiquette et pour former un modèle prédictif. Le choix des exemples à étiqueter peut être vu comme un dilemme entre l'exploration et l'exploitation sur l'espace d'entrée. Dans \cite{bouneffouf2014contextual}, une nouvelle stratégie d'apprentissage actif gère ce compromis en modélisant le problème d'apprentissage actif comme un problème de bandit contextuel.
ils proposent un algorithme séquentiel appelé Active Thompson Sampling (ATS), qui, à chaque tour, attribue une distribution d'échantillonnage sur le clusteur, échantillonne un point de cette distribution et interroge l'oracle pour cette étiquette de point d'échantillonnage. Les auteurs de \cite{ganti2013building} proposent également un algorithme d'apprentissage actif basé sur des groupes de bandits à plusieurs bras pour le problème de la classification binaire. Ils utilisent des idées telles que des limites de confiance inférieures et une régularisation auto-concordante tirée de la littérature sur les bandits à plusieurs bras pour concevoir leur algorithme. 

\subsection{Clustering}
\cite{SublimeL18} considère le clustering collaboratif, qui est un paradigme d'apprentissage automatique concerné par l'analyse non supervisée de données complexes à vues multiples à l'aide de plusieurs algorithmes fonctionnant. Les applications bien connues du clustering collaboratif incluent le clustering à vues multiples et le clustering de données distribué, où plusieurs algorithmes échangent des informations afin de s'améliorer mutuellement. L'un des principaux problèmes du clustering collaboratif et à vues multiples est d'évaluer quelles collaborations seront bénéfiques ou préjudiciables. De nombreuses solutions ont été proposées à ce problème, et toutes concluent que, à moins que deux modèles ne soient très proches, il est difficile de prédire à l'avance le résultat d'une collaboration. Pour résoudre ce problème, les auteurs de \cite{SublimeL18} proposent un algorithme collaboratif de clustering peer to peer basé sur le principe des bandits multi-bras non stochastiques pour évaluer en temps réel quels algorithmes ou vues peuvent apporter des informations utiles.

\subsection{Apprentissage par renforcement}
 Les systèmes cyber-physiques autonomes jouent un rôle important dans nos vies. Pour s'assurer que les agents se comportent de manière alignée sur les valeurs des sociétés dans lesquelles ils opèrent, nous devons développer des techniques qui permettent à ces agents non seulement de maximiser leur récompense dans un environnement, mais aussi d'apprendre et de suivre les contraintes implicites assumées par la société. Dans \cite{noothigattu2018interpretable}, les auteurs étudient un cadre où un agent peut observer des traces de comportement de membres de la société mais n'a pas accès à l'ensemble explicite de contraintes qui donnent lieu au comportement observé. Au lieu de cela, l'apprentissage par renforcement inverse est utilisé pour apprendre de telles contraintes. C'est contraintes sont ensuite combinées avec une fonction de valeur orthogonale grâce à l'utilisation d'un orchestrateur contextuel qui choisit entre deux politiques. L'orchestrateur de bandit contextuel permet à l'agent de mélanger les politiques de manière novatrice, en prenant les meilleures actions à partir d'une politique de maximisation de la récompense.

 \subsection{Bandit pour l'apprentissage automatique: \\ Résumé et orientations futures}
 
 \begin{table}[]
\scriptsize
\caption {Bandit pour l'apprentissage automatique}
\label{tab:ML} 
\begin{tabular}{|l|r|l|l|l|}
\hline
                                  & MAB          & Non            & CMAB  & Non  \\ 
                                  &              & Station-      &       & Station- \\ 
                                  &              &  MAB     &       &  CMAB\\ 
                                  \hline
Sélection d'algorithme      &              & $\surd$                  &             &                                  \\ \hline
Optimisation des paramètres  &    $\surd$   &                           &            &                                  \\ \hline
Sélection des fonctionnalités    &    $\surd$   &       $\surd$             &            &                                  \\ \hline
Apprentissage actif        &    $\surd$   &                            &   $\surd$    &                                  \\ \hline
Clustering             &   $\surd$    &                           &            &                                  \\ \hline
RL  &  $\surd$     &       $\surd$                &    $\surd$         &                                  \\ \hline
\end{tabular}
\end{table}
Le tableau \ref{tab:ML} résume les types de problèmes de bandit utilisés pour résoudre les problèmes d'apprentissage automatique mentionnés ci-dessus. Nous voyons, par exemple, que le bandit contextuel n'a pas été utilisé dans la sélection des hyperparamètres. Cette observation pourrait indiquer une direction pour les travaux futurs, où des informations secondaires pourraient être utilisées dans la sélection des caractéristiques. En outre, le bandit non stationnaire a rarement été pris en compte dans ces situations problématiques, ce qui suggère également des extensions possibles des travaux actuels. Par exemple, le bandit contextuel non stationnaire pourrait être utile dans le cadre de sélection de caractéristiques non stationnaires, où trouver les bonnes caractéristiques dépend du temps et du contexte lorsque l'environnement ne cesse de changer. Notre principale observation est également que chaque technique ne résout qu'un seul problème d'apprentissage automatique à la fois; Ainsi, la question est de savoir si un paramètre de bandit et des algorithmes peuvent être développés pour résoudre simultanément plusieurs problèmes d'apprentissage automatique, et si le transfert et l'apprentissage continu peuvent être réalisés dans ce contexte. Une solution pourrait être de modéliser tous ces problèmes dans un cadre de bandit combinatoire, où l'algorithme de bandit trouverait la solution optimale pour chaque problème à chaque itération; ainsi, le bandit combinatoire pourrait en outre être utilisé comme outil pour faire progresser l'apprentissage automatique. 
 
 \section{Conclusions}
\label{sec:Conclusion}
Dans cet article, nous avons passé en revue certains des travaux récents les plus notables sur les applications du bandit et du bandit contextuel, à la fois dans des domaines réels et dans l'apprentissage automatique. Nous avons résumé, de manière organisée (tableaux 1 et 2), diverses applications existantes, par types de paramètres de bandit utilisés, et discuté des avantages de l'utilisation des techniques de bandit dans chaque domaine. Nous décrivons brièvement plusieurs problèmes importants et des extensions futures prometteuses.
En résumé, le cadre du bandit, comprenant à la fois le bandit multi-bras et le bandit contextuel, est actuellement des domaines de recherche très actifs et prometteurs, et de multiples nouvelles techniques et applications émergent chaque année. Nous espérons que notre enquête pourra aider le lecteur à mieux comprendre certains aspects clés de ce domaine passionnant et à avoir une meilleure perspective sur ses avancées notables et ses promesses futures.

\bibliographystyle{ieeetr}
\bibliography{IEEEexample}

\end{document}